\documentclass{article}
\usepackage[final,sglblindworkshop]{neurips_2026}
\workshoptitle{Machine Learning for Systems}
\setcitestyle{numbers,square,comma}
\usepackage[T1]{fontenc}
\usepackage[utf8]{inputenc}
\usepackage{microtype}
\usepackage{booktabs}
\usepackage{graphicx}
\usepackage{amsmath}
\usepackage{amssymb}
\usepackage{tabularx}
\usepackage{array}
\usepackage{xcolor}
\usepackage[colorlinks=true,linkcolor=black,citecolor=black,urlcolor=blue!60!black,breaklinks=true]{hyperref}
\providecommand{\doi}[1]{\href{https://doi.org/#1}{\path{doi:#1}}}

\newcolumntype{Y}{>{\raggedright\arraybackslash}X}

\title{\textsc{InferOpt}: Constrained Multi-Objective Search for LLM Inference Configurations}
\author{Qi Chen \quad Yingying Cheng \quad Zhaoyi Sun \quad Li Zhou \quad Fan Zhang \quad Jie Sun \\
Huawei Technologies Co., Ltd. \\
\texttt{\scriptsize\{chenqi259, cheng.yingying1, sun.zhaoyi1, zhouli107, zhang.fan2, j.sun\}@huawei.com}}

\begin{document}
\maketitle

\begin{abstract}
Serving an LLM means setting dozens of inference-time knobs, from per-layer KV retention to per-layer expert counts. Practice sets them with mechanism-specific heuristics that return a single operating point and do not scale to layer-wise search spaces. We recast inference configuration as constrained multi-objective black-box optimization and build \textsc{InferOpt}, a reusable search framework that requires only variable bounds, a deterministic resource cost, and an evaluation hook. \textsc{InferOpt} searches on a frozen sampled proxy set, rejects over-budget candidates before any model call, tightens the budget adaptively, and re-validates Pareto representatives
on full-scale dataset. One pipeline covers a 28-dimensional continuous KV
space (Qwen2.5-7B) and a 26-dimensional discrete MoE space (DeepSeek-V2-Lite).
On KV, post-prefill pruning cuts the 16K cache by 64.4\% and TPOT by
22.9--48.5\%, and the searched layer-wise budget by \textsc{InferOpt} beats a matched uniform budget
by 7.3\% and 14.0\% of the Full KV reference points. On MoE, a searched top-$k$ schedule by \textsc{InferOpt} removes 43.0\% of routed
token--expert pairs while staying within 0.59 points of the default, closer than the matched-budget baselines. Against
Random Search, NSGA-II, and MOTPE, \textsc{InferOpt} leads on both spaces,
taking the best proxy hypervolume and the lowest retention on KV and staying
closest to the uncompressed reference at the lowest experts on MoE.
\end{abstract}

\section{Introduction}

LLM serving systems expose configuration choices for KV-cache compression \citep{liu2024kivi}, sparse attention, MoE routing \citep{lu2024expertpruning}, and quantization, which are typically determined by uniform policies or mechanism-specific importance heuristics \citep{he2024demystifying}. With dozens of layer-wise variables, these heuristics become inadequate. They explore the high-dimensional configuration space only narrowly and yield a single operating point rather than the quality-efficient Pareto frontier required under varying deployment budgets.

We argue the missing piece is not another compression heuristic but an
\emph{optimization layer}. InferOpt represents continuous KV-memory fractions and discrete routing decisions within a unified mixed-domain configuration space. Supporting a new mechanism requires only the specification of its variable domains, resource-cost function, and evaluation interface. The same search and validation pipeline is reused without mechanism-specific redesign. Evaluation cost is the primary bottleneck. Evaluating a single configuration of a billion-parameter model on the full benchmark can take hours, making conventional multi-objective search impractical under realistic budgets and motivating a sample-efficient search framework.

\textsc{InferOpt} addresses these challenges through three empirically validated components.
\textbf{(i) Frozen proxy evaluation with explicit reliability auditing.} A fixed proxy set ranks candidate configurations during search, after which selected configurations undergo full-benchmark validation.
\textbf{(ii) Deterministic resource-aware pruning.} A deterministic resource-cost function rejects over-budget candidates before model evaluation and progressively tightens the admissible budget as feasible observations accumulate.
\textbf{(iii) A mechanism-agnostic search interface.} By exposing only variables, constraints, and objectives, the same search and validation pipeline supports both continuous KV-cache and discrete MoE-routing spaces without mechanism-specific redesign. Across both spaces, \textsc{InferOpt} outperforms Random Search, NSGA-II, and MOTPE.

\paragraph{Related work.}
Existing inference-acceleration methods \citep{liu2024kivi,li2024snapkv,zhou2022expertchoice} introduce effective configuration mechanisms, but typically optimize them in isolation. General-purpose multi-objective optimization methods \citep{bergstra2011tpe,akiba2019optuna,shahriari2016review} support heterogeneous search spaces, but rely on relatively inexpensive objective evaluations. \textsc{InferOpt} bridges this gap by extending the abstraction of ML auto-tuners \citep{chen2018learning,zheng2020ansor} from low-level tensor schedules to expensive, high-level inference configurations.

\section{Method}
Let $\mathbf{x}\in\mathcal{X}$ collect per-layer retention ratios, expert
counts, or any other inference configuration. We consider the following multi-objective optimization problem including resource cost and quality loss,
\begin{equation}
\min_{\mathbf{x}\in\mathcal{X}}
\bigl(\mathrm{ResourceCost}(\mathbf{x}),\;Q_{\rm ref}-Q(\mathbf{x},\mathcal{D})\bigr)
\quad\text{s.t.}\quad g_j(\mathbf{x})\leq 0,
\label{eq:moo}
\end{equation}
where $Q_{\rm ref}$ is the score of the uncompressed reference on the same
examples. Resource cost is an architecture-level proxy computed in closed form,
namely mean retained historical KV blocks or mean activated routed experts, not
measured latency, memory, or energy.

\begin{figure}[!ht]
  \centering
  \includegraphics[width=1.0\linewidth]{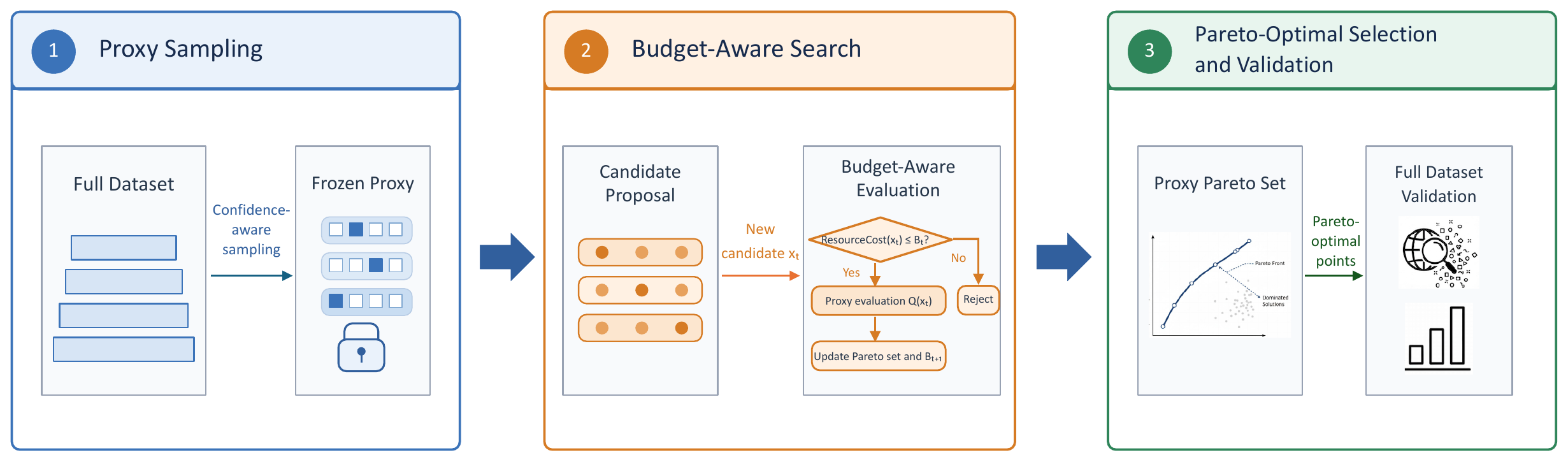}
  \caption{\textsc{InferOpt} in three stages. Proxy sampling constructs a frozen
  stratified subset; budget-aware multi-objective search rejects over-budget
  candidates before evaluation; Pareto-optimal representatives are selected and
  re-validated on disjoint held-out examples.}
  \label{fig:overview_ml4sys}
\end{figure}

Figure~\ref{fig:overview_ml4sys} illustrates the three-stage
\textsc{InferOpt} pipeline.
\textbf{Proxy sampling.}
We first construct a proxy subset from the full dataset using confidence-aware
sampling and freeze it before search. Consequently, every candidate is evaluated
on the same examples.
\textbf{Budget-aware search.}
At iteration $t$, multi-objective TPE proposes a continuous or discrete
configuration $\mathbf{x}_t$ based on the Pareto structure of previously
evaluated trials \citep{bergstra2011tpe}. Before model
evaluation, a closed-form resource-cost function checks whether
$\mathrm{ResourceCost}(\mathbf{x}_t) \leq B_t$. Over-budget candidates are
rejected immediately, whereas feasible candidates are evaluated on the frozen
proxy set and used to update both the proxy Pareto set and the next-round budget
$B_{t+1}$.
\textbf{Pareto-optimal selection and validation.}
Finally, Pareto-optimal configurations are selected from the proxy Pareto set and re-evaluated on the full dataset to identify deployment-ready operating points.

\paragraph{Two instantiations.}
We instantiate \textsc{InferOpt} on two structurally different configuration
spaces.
\textbf{KV-cache compression.}
The configuration is a 28-dimensional continuous vector
$\boldsymbol{\rho}\in[0.1,1.0]^{28}$, where $\rho_\ell$ denotes the retention
ratio of layer $\ell$. We define the resource cost as the average retention
ratio,
$\bar{\rho}=\frac{1}{28}\sum_{\ell=1}^{28}\rho_\ell$,
and impose the adaptive constraint $\bar{\rho}\leq B_t$, where $B_0=1.0$ and
$B_t$ is lower-bounded by $0.2$.
\textbf{MoE routing.}
The configuration is a 26-dimensional discrete vector
$\mathbf{k}\in\{1,\ldots,6\}^{26}$, where $k_\ell$ denotes the number of
activated experts at layer $\ell$. Its resource cost is
$\bar{k}=\frac{1}{26}\sum_{\ell=1}^{26}k_\ell$, with no hard constraint imposed
on the total number of activated experts during search.
Apart from the variable domains and resource-cost definitions, the search and
validation pipeline remains unchanged across the two instantiations.

\section{Experiments}
\paragraph{Setup.}
All runs use one GPU, Python 3.11, PyTorch 2.5.1, and bfloat16. KV compression
uses Qwen2.5-7B-Instruct \citep{qwen2024qwen25} on LongBench
\citep{bai2024longbench}, with 210 proxy and 1,050 held-out examples over 21
tasks and a protected 128-token prefix and 512-token recent window. MoE routing
uses DeepSeek-V2-Lite-Chat \citep{deepseekai2024v2} with 64 experts per routed
layer, on 384 proxy and 768 held-out MMLU/GSM8K/BBH
examples \citep{hendrycks2021mmlu,cobbe2021gsm8k,suzgun2023bbh} and checkpoint
default $k_\ell=6$. Multi-objective search baselines are Random Search, NSGA-II \citep{deb2002fast},
and MOTPE \citep{bergstra2011tpe,akiba2019optuna} under the same box
constraints but without adaptive budget-tightening. Appendix~\ref{app:protocol} gives the
remaining protocol details.

\begin{table}[!htbp]
\centering\scriptsize
\caption{Comparison of multi-objective search methods. Left: KV proxy search over 24 trials, hypervolume against the reference point on the
210-example proxy. The last column is quality loss
in points at the minimum-retention trial, where full KV scores 40.25/100. Right: MoE 24-trial replay on the 384-example proxy, max and mean score comes from six selected candidates per method. The uncompressed
default $k_\ell=6$ reference scores 0.1953 (19.53/100). MoE scores are read by proximity to that
reference rather than by magnitude. Bold marks the best value in a column.}
\label{tab:search_methods}
\begin{minipage}[t]{0.53\linewidth}
\centering
\begin{tabular}{@{}lccc@{}}
\toprule
Method & Hypervolume & Min. $\bar\rho$ & Loss at min. $\bar\rho$ \\
\midrule
Random   & $0.00788$ & $0.313$ & $0.166$ \\
NSGA-II  & $0.00795$ & $0.331$ & $-0.033$ \\
MOTPE    & $0.00778$ & $0.327$ & $0.005$ \\
\textsc{InferOpt} & $\mathbf{0.00881}$ & $\mathbf{0.274}$ & $0.053$ \\
\bottomrule
\end{tabular}
\end{minipage}%
\hfill
\begin{minipage}[t]{0.45\linewidth}
\centering
\begin{tabular}{@{}lcccc@{}}
\toprule
Method & Max & Mean & $\sum_\ell k_\ell$ & $\bar{k}$ \\
\midrule
Random   & $0.2240$ & $0.1836$ & $65.7$ & $2.53$ \\
NSGA-II  & $0.2179$ & $0.1832$ & $95.3$ & $3.67$ \\
MOTPE    & $0.2240$ & $0.1840$ & $75.7$ & $2.91$ \\
\textsc{InferOpt} & $\mathbf{0.1931}$ & $\mathbf{0.1910}$ & $\mathbf{59.3}$ & $2.28$ \\
\bottomrule
\end{tabular}
\end{minipage}
\end{table}

\paragraph{Search-method comparison.}
Table~\ref{tab:search_methods} compares the four multi-objective search methods. On
the KV proxy, \textsc{InferOpt} attains the highest mean hypervolume ($0.00881$,
$10.8\%$ above the best baseline) and the lowest minimum retention ($0.274$,
$12.5\%$ below it) at a quality loss of $0.053$ points ($0.1\%$ of the Full KV
proxy reference). On the MoE panel, \textsc{InferOpt} leads all
three columns. Its top candidate scores $0.1931$, within $0.22$ points of the
reference ($1.1\%$), against $11.6$--$14.7\%$ for the three baselines; its
selected set averages $0.1910$, again the closest at $2.2\%$; and it reaches
both at the lowest budget, $\sum_\ell k_\ell=59.3$, $9.7$--$37.8\%$ fewer
activated experts than the baselines.

\subsection{KV cache compression}

\textbf{Controlled comparison.}
We fix an aggressive mean-retention budget ($\bar{\rho}=0.321$) and evict
unselected KV blocks immediately after prefill. We independently vary
\emph{allocation} and \emph{selection}: allocation uses either the layer-wise
vector found by \textsc{InferOpt} or a matched uniform vector, while selection
uses attention-mass scoring \citep{li2024snapkv}, key-$\ell_2$ magnitude, or an
H2O-style policy \citep{zhang2023h2o}. All five compressed policies share the
same protected regions and mean budget, enabling controlled comparisons of the
allocation and selection effects.

\begin{figure}[!ht]
  \centering
  \begin{minipage}[t]{0.315\linewidth}
    \centering
    \includegraphics[width=\linewidth]{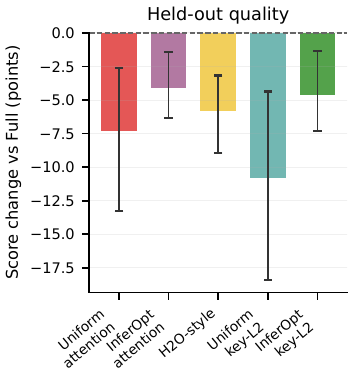}
    \par\vspace{1pt}\makebox[\linewidth]{\small (a)}
  \end{minipage}%
  \hfill
  \begin{minipage}[t]{0.315\linewidth}
    \centering
    \includegraphics[width=\linewidth]{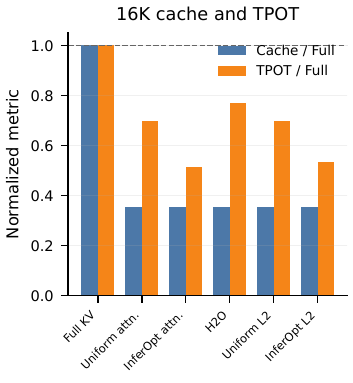}
    \par\vspace{1pt}\makebox[\linewidth]{\small (b)}
  \end{minipage}%
  \hfill
  \begin{minipage}[t]{0.315\linewidth}
    \centering
    \includegraphics[width=\linewidth]{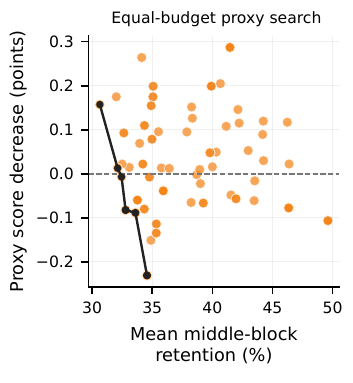}
    \par\vspace{1pt}\makebox[\linewidth]{\small (c)}
  \end{minipage}
  \caption{KV results at a fixed aggressive budget $\bar\rho=0.321$. (a)
  Validation/Held-out quality change against Full KV on 1,050 paired examples; Full KV scores 43.83/100. (b)
  Same-process 16K cache and TPOT, normalized to Full KV over ten trials.
  (c) Every KV compression candidate evaluated during the proxy search of InferOpt, black curve marking
  the non-dominated front.}
  \label{fig:kv_results_pair}
\end{figure}

\textbf{Quality.}
Figure~\ref{fig:kv_results_pair}(a) shows that all compressed policies trade
quality for efficiency, decreasing the Full-KV score of $43.83$ by
$5.79$--$10.78$ points on 1,050 validation examples. The
comparisons show that allocation has the larger effect. With allocation fixed,
attention-mass scoring outperforms key-$\ell_2$ by only $0.52$ points. With the
selector fixed, the searched allocation outperforms its budget-matched uniform
counterpart by $3.21$ points with attention scoring and $6.13$ points with
key-$\ell_2$.

\textbf{Efficiency and search behavior.}
At a 16K context length, panel~(b) shows that \textsc{InferOpt} reduces KV-cache
memory from $902.9$ to $319.1$--$321.5$ MiB ($64.4\%$) and TPOT from $19.11$ ms
to $9.83$ ms with attention scoring and $10.22$ ms with key-$\ell_2$. These
latencies are lower than those of the matched uniform policies
($13.34$--$13.36$ ms) and the H2O-style baseline ($14.74$ ms). Panel~(c)
visualizes the underlying search: each orange point denotes a candidate
evaluated, and the black curve traces the resulting
proxy Pareto front. At nearly identical cache budgets, the selected Pareto-frontal
configurations outperform the matched uniform baselines by $3.21$ and $6.13$ (7.3\% and 14.0\%) 
quality points and reduce TPOT by $3.5$ and $3.1$ ms (26.4\% and 23.4\%), respectively. Thus,
\textsc{InferOpt} exposes the full quality--efficiency frontier.

\subsection{\texorpdfstring{MoE top-$k$ routing}{MoE top-k routing}}

Table~\ref{tab:moe_short} compares five layer-wise top-$k$ schedules on a
768-example validation set: the \textsc{InferOpt} schedule, the uncompressed
default, two budget-matched controls, and a lower-budget uniform endpoint.
Routed pairs counts
token--expert assignments, and TPOT is measured at an 8K context length.
Score denotes macro-averaged validation quality, while $\Delta$ measures its
absolute deviation from the uncompressed default. The selected schedule,
\texttt{seg\_early6\_mid3\_late1}, assigns $k=6,3,1$ to the first 9, middle 9,
and final 8 layers, respectively, yielding $\bar{k}=3.42$. It achieves the
smallest quality deviation, remaining within
$0.59$ points ($3.3\%$) of the default, while
reducing routed pairs by $43.0\%$ and TPOT by $19.3\%$. These results indicate that preserving expert capacity in early
layers while aggressively pruning later layers provides a favorable
quality-efficiency trade-off.

Two matched-budget controls test that shape at $\sum_\ell k_\ell=89$.
Near-uniform spreads the 11 extra experts over evenly spaced layers;
entropy-based allocation gives them to the 11 layers with the highest
normalized entropy of default-$k$ activation counts on the proxy. The segmented schedule lands
0.45 points (2.5\% of the default score) closer to the default than
near-uniform and 0.97 points (5.4\%) closer than entropy-based. We built both
controls only after the winning schedule was known, so they are a post-hoc
check rather than a pre-registered test. The
three schedules spend the same budget and differ only in which layers get the
extra experts, so the gap between them cannot come from budget, and their TPOT
differences are runtime effects for the same reason.

\begin{table}[!htbp]
\centering
\caption{MoE quality and 8K system measurements over full validation set. The two
matched-budget controls are post-hoc; uniform $k=3$ is a lower-budget endpoint
rather than a matched control.}
\label{tab:moe_short}
\begin{tabular}{@{}lrrrrr@{}}
\toprule
Configuration & Mean $k$ & Score & $\Delta$ & Routed pairs & TPOT (ms) \\
\midrule
\texttt{seg\_early6\_mid3\_late1} & 3.42 & 0.1725 & $-0.0059$ & 17,406 & 32.46 \\
Default $k=6$ & 6.00 & 0.1784 & $0.0000$ & 30,510 & 40.23 \\
Near-uniform, $\sum k_\ell=89$ & 3.42 & 0.1680 & $-0.0104$ & 17,406 & 35.73 \\
Entropy-based, $\sum k_\ell=89$ & 3.42 & 0.1628 & $-0.0156$ & 17,406 & 35.72 \\
Uniform $k=3$ & 3.00 & 0.1589 & $-0.0195$ & 15,255 & 32.26 \\
\bottomrule
\end{tabular}
\end{table}

\section{Conclusion}

We introduced \textsc{InferOpt}, a unified constrained multi-objective search
framework in which each inference mechanism specifies only its variable
domains, resource-cost function, and evaluation interface. The same pipeline
supports continuous KV-retention and discrete MoE-routing spaces without
mechanism-specific redesign and outperforms Random Search, NSGA-II, and MOTPE
in both settings. The searched KV allocation improves
quality by $3.21$--$6.13$ points (7.3\% and 14.0\%) and reduces TPOT by $3.1$--$3.5$ ms (26.4\% and 23.4\%) over
uniform allocation, while using $64.4\%$ less cache than Full KV at 16K. The
searched MoE schedule reduces routed work by $43.0\%$ while remaining within
$0.59$ points of the default validation score. These results demonstrate that
a shared, budget-aware search framework can effectively replace
mechanism-specific tuning heuristics.

{\small
\bibliographystyle{plainnat}
\bibliography{references}
}

\appendix
\section{Protocol details}
\label{app:protocol}
\paragraph{KV compression.}
Every compressed policy uses \texttt{prefill\_after} compression with a
128-token protected prefix, a 512-token recent window, 128 middle blocks, a
32-step compression interval, and 128-token prefill attention query chunks, at
a fixed mean middle-block retention $\bar\rho=0.321$. Compressed policies prune
once immediately after prefill and then at the compression interval during
decoding, while Full KV is a generation-only baseline. Held-out quality is the
macro-average over 21 LongBench tasks on 1,050 paired examples, with intervals
from 1,000 paired bootstrap resamples. The system matrix covers six strategies
at 4K, 8K, and 16K context with one warm-up run, ten measured repetitions, and
128 generated tokens per run; the main text reports the 16K endpoint. Because
the five compressed policies share protected regions and mean budget, uniform
attention, uniform key-L2, and H2O-style retain the same block count and the
two \textsc{InferOpt} policies share one layer-wise vector; the 321.5 versus
319.1 MiB gap is per-layer block rounding.

\paragraph{MoE routing.}
The proxy objective is the macro-average over the 384-example MMLU/GSM8K/BBH
split; a 0.02 quality tolerance labels proxy candidates but is not a hard
feasibility constraint. The proxy-to-held-out Kendall correlation is
$\tau_b=0.59$ with 80\% of proxy-front points preserved, so the proxy supports
coarse ranking and is no substitute for the held-out re-scoring every reported
schedule receives. Routed token--expert pairs are an architecture-level
work proxy and are identical across the three $\sum_\ell k_\ell=89$
allocations, so uniform $k=3$ in Table~\ref{tab:moe_short} is a lower-budget
endpoint rather than a matched control.

\end{document}